\documentclass[sigconf,nonacm]{acmart}
\renewcommand\footnotetextcopyrightpermission[1]{}

\usepackage{booktabs} % For formal tables

\usepackage{subfigure}
\usepackage{algorithm}
\usepackage{algorithmic}
\usepackage{multirow}
\usepackage{amsmath}

\acmDOI{10.475/123_4}

\acmISBN{123-4567-24-567/08/06}

\begin{document}

\title{Attribute-Induced Bias Eliminating for Transductive Zero-Shot Learning}
\subtitle{Hantao Yao, Shaobo Min, Yongdong Zhang, Changsheng Xu\\National Laboratory of Pattern Recognition, CASIA\&USTC}

\begin{abstract}
Transductive Zero-shot learning (ZSL) targets to recognize the unseen categories by aligning the visual and semantic information in a joint embedding space.
There exist four kinds of domain biases in Transductive ZSL, \emph{i.e.,} ~\emph{visual bias} and ~\emph{semantic bias} between two domains and two ~\emph{visual-semantic biases} in respective seen and unseen domains, but existing work only focuses on the part of them, which leads to severe semantic ambiguity during the knowledge transfer.
To solve the above problem, we propose a novel Attribute-Induced Bias Eliminating (AIBE) module for Transductive ZSL.
Specifically, for the \emph{visual bias} between two domains, the Mean-Teacher module is first leveraged to bridge the visual representation discrepancy between two domains with unsupervised learning and unlabelled images.
Then, an attentional graph attribute embedding is proposed to reduce the \emph{semantic bias} between seen and unseen categories, which utilizes the graph operation to capture the semantic relationship between categories.
Besides, to reduce the semantic-visual bias in the seen domain, we align the visual center of each category, instead of the individual visual data point, with the corresponding semantic attributes, which further preserves the semantic relationship in the embedding space. 
Finally, for the semantic-visual bias in the unseen domain, an unseen semantic alignment constraint is designed to align visual and semantic space in an unsupervised manner.
The evaluations on several benchmarks demonstrate the effectiveness of the proposed method,  \emph{e.g.,} obtaining the 82.8\%/75.5\%, 97.1\%/82.5\%, and 73.2\%/52.1\% for Conventional/Generalized ZSL settings for CUB, AwA2, and SUN datasets, respectively.
\end{abstract}



%\ccsdesc[500]{Computer systems organization~Embedded systems}

% We no longer use \terms command


\maketitle

\section{Introduction}
Given the image labels for seen domain and semantic description for all categories, the zero-shot learning (ZSL) aims to recognize the objects belonging to unseen categories. 
Based on the assumption that the seen and unseen domains share the common semantic and visual spaces, constructing the relationship between two modalities can facilitate the knowledge from seen to unseen domains, thereby enabling the model to recognize the unseen objects.
Since ZSL can release the dependency in the unseen domain, it has been attracted much attention recently~\cite{norouzi2013zero,jia2019deep,wang2019survey,fu2015transductive,snell2017prototypical,kampffmeyer2019rethinking,song2018selective,romera2015embarrassingly,rohrbach2013transfer}.

\begin{figure}
	\begin{center}
		\includegraphics[width=0.9\linewidth]{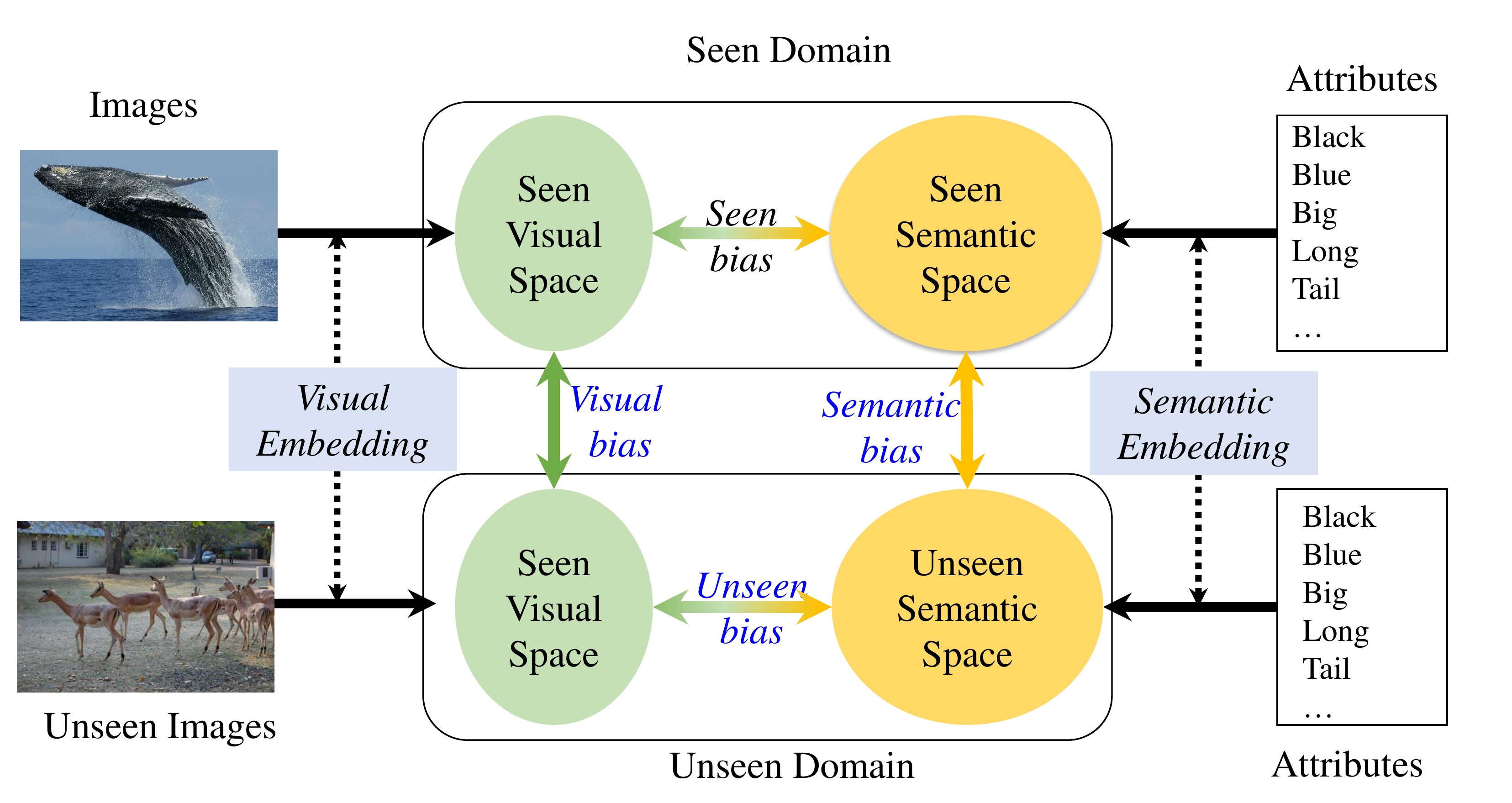}
	\end{center}
	\caption{The bias existed in Transductive Zero-shot learning (ZSL), \emph{e.g.,} \emph{visual bias, semantic bias, seen bias, and unseen bias.} Besides the \emph{visual bias}, this work also focuses on associating the visual and semantic space for unseen domains by reducing the other three biases that are all related to the unseen domain.}
	\label{fig1}
\end{figure}

Based on whether the unseen images are accessible during training, zero-shot learning can be divided into \emph{Inductive zero-shot learning (ZSL)}~\cite{xian2018zero,li2018discriminative,hubert2017learning, zhu2019generalized,sariyildiz2019gradient,changpinyo2020classifier} and \emph{Transductive zero-shot learning (ZSL)}~\cite{fu2015transductive, song2018transductive,wan2019transductive,rohrbach2013transfer}.
In Inductive ZSL, since the unseen images are unavailable during training, it suffers from the serious visual bias between two domains \cite{zhang2017learning,Annadani2018}.
Differently, the images and semantic attributes for the unseen domain can be accessed in Transductive ZSL, but the relationship between images and semantic attributes is unavailable, which is a more realistic scene than the Inductive ZSL. 
A general paradigm of ZSL is to learn a common visual-semantic embedding space based on the seen domain, which alleviates the semantic-visual bias,~\emph{i.e.,} \emph{seen bias}, as shown in Figure~\ref{fig1}.
Since the seen and unseen domains share a common visual and semantic space, such a semantic-visual knowledge can be directly transferred to the unseen domain \cite{xian2018zero,li2018discriminative}.
However, the discrepancies between the visual and semantic representations of two domains significantly impede the generalization to the unseen domain.

Based on the provided images and semantic attributes for seen and unseen domains, there exactly exist four types of domain biases in Transductive ZSL, as shown in Figure~\ref{fig1}: \emph{seen bias} and \emph{unseen bias} denote the discrepancy between visual and semantic information for seen and unseen domains, respectively; \emph{visual bias} and \emph{semantic bias} denote the visual and semantic distribution shift between seen and unseen domains, respectively.
To address the above issues, existing Transductive ZSL targets to utilize the provided unseen data to alleviate the visual and semantic bias problem.
For example, UDA \cite{kodirov2015unsupervised} jointly learns a regularised sparse coding for the semantic embedding learning in two domains, and ReViSE \cite{hubert2017learning} minimizes the mean discrepancy of visual embeddings between two domains via an auto-encoder architecture. 
Recently, QFSL \cite{song2018transductive} bridges the \emph{visual bias} between two domains via a quasi-fully supervised manner, which directly punishes the unseen instances being recognized as the seen classes.
Although existing approaches are effective, none of the can explicitly define and tackle all four types of biases in a unified framework.
As a consequence, they still suffer from a serious semantic ambiguity problem when transferred to the unseen domain.

Inspired by unsupervised and semi-supervised learning ~\cite{tarvainen2017mean,laine2016temporal,french2017self}, we propose a novel Attribute-Induced Bias Eliminating (AIBE) module that simultaneously alleviates the \emph{seen bias}, \emph{unseen bias}, \emph{visual bias}, and \emph{semantic bias} problems in Transductive ZSL.
The core motivation of AIBE is first to leverage the Mean-Teacher and graph network \cite{kipf2016semi} during visual and semantic embedding, respectively, to preserve the consistent visual distribution and semantic relationship between two domains.
Then, a consistency constraint is further designed to align the visual and semantic spaces of the unseen domain with an unsupervised manner.

Based on the above discussion, the AIBE consists of four sub-modules.
First, AIBE treats the visual center, which is the mean of all visual descriptions belonging to each category, as the supervision information to align the semantic space with visual space for the seen domain,~\emph{i.e.,} ~\emph{seen bias}.
Then, with the labeled seen images and unlabeled unseen images, AIBE applies the Mean-Teacher (MT)~\cite{tarvainen2017mean} to infer a discriminative visual space for both seen and unseen domains, which bridges the ~\emph{visual bias}.
Besides, for the \emph{semantic bias}, an Attentional Graph Attribute Embedding (AGAE) is proposed to model the attribute relationship between all categories, which can preserve the subtle semantic topology in the joint embedding space.
Finally, an Unseen Semantic Alignment (USA) constraint is proposed to reduce the bias between visual and semantic spaces for the unseen domain.
After obtaining the visual embedding and semantic attribute embedding, the Nearest Neighbor (NN) search is applied to recognize the testing objects.

The evaluation on three benchmarks demonstrate the effectiveness of the proposed method,  \emph{e.g.,} obtaining the 82.8\%/75.5\%, 97.1\%/82.5\%, and 73.2\%/52.1\% for Conventional/Generalized ZSL settings for CUB, AwA2, and SUN datasets, respectively.
Our contributions can be summarized as follows:
\begin{enumerate}
\item By considering the unlabeled unseen images provided in Transductive ZSL, we find out that using unsupervised learning for unseen images can boost the visual representation for the unseen domain, and can reduce the \emph{visual bias}.
\item We demonstrate that considering the relationship of the semantic attribute can reduce the semantic gap between seen and unseen domains.
\item We prove that constructing an unsupervised constraint between visual and semantic spaces for the unseen domain is essential to reduce the bias between visual and semantic spaces for the unseen domain.
\end{enumerate}

\section{Related Work}
Most of the Zero-Shot Learning (ZSL) methods are designed under the Transductive and Inductive manners, thus we review these two kinds of approaches.
\subsection{Transductive Zero-Shot Learning}
In the Transductive ZSL, the images and categories of the unseen domain are available during training, but the corresponding image-category relationship is unknown.
Thus, many methods focus on utilizing the unseen domain instances to minimize the visual and semantic discrepancy between two domains.
For example, Fu~\emph{et.al.,}~\cite{fu2015transductive} leverage the test instances and CCF to learn a hyper-graph among two domain semantics in the joint embedding space.
UDA \cite{kodirov2015unsupervised} jointly learns a regularised sparse coding to associate two domain samples, which effectively alleviates the biased recognition problem. and ReViSE \cite{hubert2017learning} minimizes the mean discrepancy between two domain embeddings via an auto-encoder architecture. 
Recently, QFSL \cite{song2018transductive} utilizes the unseen instances to bridge the visual bias between two domains via a quasi-fully supervised manner.
Specifically, QFSL applies a strong penalty to the situation where the unseen instances are recognized as the seen categories. 
By designing elaborate constraints, these methods explore the unseen instances to successfully bridge either visual or semantic domain bias problem and obtain the promising results.

Besides domain discrepancy minimizing, some methods target to generate pseudo-labels for the unseen instances via label propagation, and then learn the classifier in a fully supervised manner.
In \cite{guo2017zero}, a quadratic formulation is proposed to generate pseudo-labels for the unseen instances, which considers both the reliability and diversity of the unseen samples simultaneously, and then an unseen classifier is trained in the original feature space with the pseudo-labels.
Similarly, PREN \cite{ye2019progressive} constructs multi-classifiers for the label embeddings from different domains, which enables the information transfer from the seen categories to the unseen categories and generate the pseudo-labels through adaptive inter-label relations.
Recently, the generation-based methods,~\emph{e.g.,} GAN, obtain the state-of-the-art performance, which directly synthesize the unseen data-label pairs from category descriptions to train a fully-supervised model.
For example, Verma~\emph{et.al.,}~\cite{verma2017simple} develop a generative probability model to  represent each class naturally, and SABR \cite{paul2019semantically} leverages Wasserstein GAN to synthesize the unseen domain distribution, which is defined by the available unseen instance.

The proposed method of this paper belongs to Transductive ZSL, and the main difference from the previous methods is that we simultaneously consider four kinds of bias problem,~\emph{i.e.,} seen modality bias, unseen modality bias, visual domain bias, and semantic domain bias, while the other methods only tackle some of them. 

\begin{figure*}
	\begin{center}
		\includegraphics[width=0.8\linewidth]{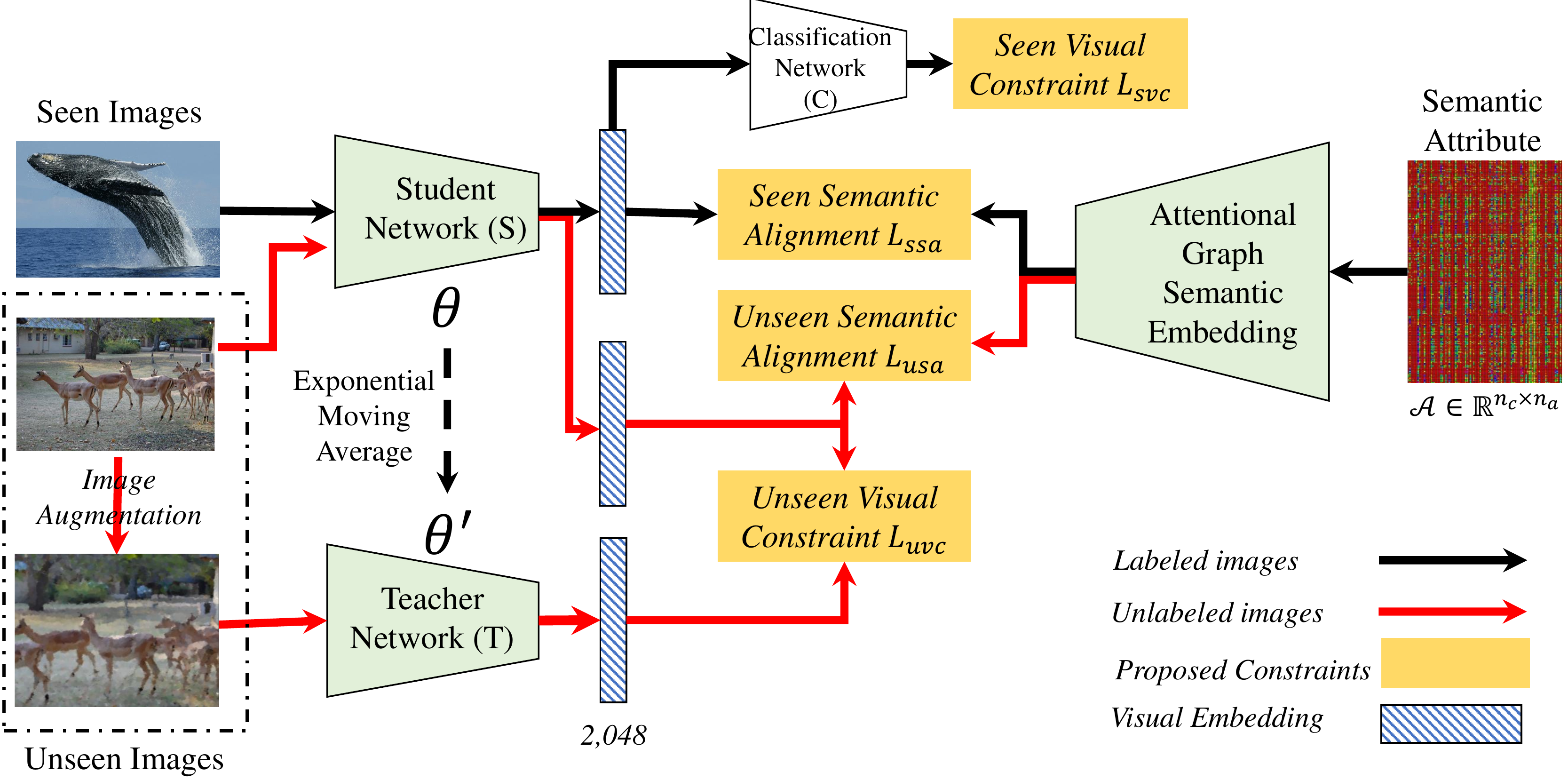}
	\end{center}
	\caption{The framework of the Attribute-Induced Bias Eliminating. The proposed module consists of two major components: visual embedding and semantic attribute embedding. The visual embedding is implemented based on the Mean-Teacher module. The labeled seen images are fed into the Student Network (S) and Classification Network (C) for representation learning with \emph{Seen Visual Constraint $L_{svc}$}. Unlabeled unseen images along with its augmentations are fed into the Student Network (S) and Teacher Network (T) for feature extraction, respectively, and consistency loss is applied, \emph{i.e.,} \emph{Unseen Visual Constraints} , for unseen visual embedding. Given the attributes for all categories, the Attentional Graph Semantic Embedding (AGSM) is proposed for semantic embedding, and two constraints are also proposed to align the visual and semantic spaces.}
	\label{fig2}
\end{figure*}

\subsection{Inductive Zero-Shot Learning}
Different from the Transductive ZSL, the Inductive ZSL is a more strict case, when the unseen instance is also unavailable.
Thus, existing Inductive ZSL methods design kinds of joint embedding space, where the visual images and semantic labels are aligned, and transfer the embedding to the unseen domain.
The early works \cite{xian2018zero,li2018discriminative,hubert2017learning} directly utilize the semantic space spanned by the category description as the embedding space, which projects the visual feature into semantic labels.
However, due to the small dimension of semantic space, these methods suffer from the Hubness problem \cite{Radovanovic2010,Tomasev2014},~\emph{i.e.,} a few samples becoming the cluster center for the most of the query points.
Thus, recent methods \cite{zhang2017learning,Annadani2018, zhang2020towards} use the visual space spanned by high-dimensional visual features as the embedding space, which obtains new state-of-the-art performance, and design some rules to preserve the semantic topology in the embedding space.
Besides, some methods learn an intermediate space between the visual features and semantic labels , which takes both advantages of visual and semantic embedding methods but is also much harder to learn. 

\section{Attribute-Induced Bias Eliminating}
\subsection{Problem Formulation}
Transductive Zero-shot learning (ZSL) aims to recognize objects belonging to the unseen categories with the help of the labeled images and semantic category attributes.
Formally, defining the datasets as $\mathcal{D}=\{\mathcal{D}_{s}, \mathcal{D}_{u}, \mathcal{A}\}$, where $\mathcal{D}_{s}=\{\mathcal{X}_{s}, \mathcal{Y}_{s}, \mathcal{A}_{s}\}$ are the training sets,  and $\mathcal{D}_{u}=\{\mathcal{X}_{u}, \mathcal{Y}_{u}, \mathcal{A}_{u}\}$ are test sets.
$\mathcal{X}$ and $\mathcal{Y}$ denote the images and corresponding labels, respectively.
$\mathcal{A}=\mathcal{A}_{s}\cap\mathcal{A}_{u}$ denotes the semantic information for all categories, where $\mathcal{A}_{s}$ and $\mathcal{A}_{u}$ are the semantic attributes for seen and unseen categories, respectively. 
Based on the assumption that the unseen classes and seen classes are disjoint, ~\emph{i.e.,} $\mathcal{Y}_{s}\cup\mathcal{Y}_{u}=\oslash$, Transductive ZSL is to infer the category for the input image from $\mathcal{X}_{u}$ with $\mathcal{D}_{s}$ and $\mathcal{A}$.

In the Transductive ZSL, there exist four types of embedding spaces, \emph{i.e.,} seen visual space, unseen visual space, seen semantic space, and unseen semantic space, as shown in Figure~\ref{fig1}.
Since the seen and unseen domains share the same visual and semantic spaces, the critical problem of Transductive ZSL becomes how to associate the unseen visual and unseen semantic spaces.
For the seen visual space, we learn a projection function to associate the visual representation and semantic embedding with reducing the \emph{seen bias}, which denotes the bias between visual and semantic spaces for the seen domain.
However, when transferring the projection function to the unseen domain, there exist inevitable distribution biases of visual representation and semantic information between two domains,~\emph{i.e.,} \emph{visual bias} and \emph{semantic bias}, which seriously impede the knowledge transfer in zero-shot learning.

To overcome the above issue, by considering the unlabeled unseen images,  we apply the Mean-Teacher model to infer a joint visual space for both seen and unseen categories to reduce the \emph{visual bias}.
Once obtaining the unbiased visual embedding, the semantic embedding is the other critical factor for Transductive ZSL.
An ideal semantic embedding should satisfy the following two conditions: 1) the seen and unseen semantic spaces have a small bias; 2) the semantic space and visual space for the unseen domain are alignments.
Existing semantic embedding apporaches~\cite{xian2018zero,li2018discriminative,hubert2017learning} often embed semantic attributes by reducing the distance between the visual and semantic representation for each category.
The shortcoming of these works is that they cannot reduce the semantic bias between seen and unseen domains.
By considering the relationship between semantic attributes for all categories, we propose an Attentional Graph Attribute Embedding to transfer the semantic attributes to semantic space, and reduce the \emph{semantic bias}.
Further, an Unseen Semantic Alignment (USA) constraint is proposed to reduce the gap between visual and semantic space for unseen images.
The framework of the proposed module is illustrated in Figure~\ref{fig2}, and we will give a detailed description of each component in the following.
 
\subsection{Unbiased Visual Embedding}
Due to the inevitable distribution shift between seen and unseen images, the visual model inferred from the seen domain cannot obtain an unbiased visual embedding for the unseen domain. 
Therefore, we propose an Unbiased Visual Embedding model by considering the labeled images $\mathcal{X}_{s}$ and unlabelled images $\mathcal{X}_{u}$ simultaneously during visual representation learning.  
As shown in Figure~\ref{fig2}, the Unbiased Visual Embedding consists of two types of constraints: Seen Visual Constraint $L_{svc}$, and Unseen Visual Constraint $L_{uvc}$.
The seen visual constraint $L_{svc}$ is used to make the visual space discriminative enough for the seen domain, and the unseen visual constraint $L_{uvc}$ is applied to infer the unseen visual embedding.

Given the seen images $\mathcal{X}_{s}$ along with its labels $\mathcal{Y}_{s}$, and unseen images $\mathcal{X}_{u}$, we apply the unsupervised representation learning module Mean Teacher to implement the Unbiased Visual Embedding.
As shown in Figure~\ref{fig2}, the module consists of three sub-networks: student network, teacher network, and classification network.
The student and teacher networks are used to embed the input images into visual space, and the classification network is used to constrain the labeled seen images.
Noted that the teacher network has the same structure as the student network, and the weights in the teacher network are an exponential moving average of the weights in the student network.
Formally, the weights for the student network and the teacher network denote as $\theta$ and $\theta'$, respectively, where $\theta'$ is an exponential moving average of $\theta$. 
At iteration $k$, we have:
\begin{equation}
    \theta'(k) = \omega_{1}\theta'(k) + (1 - \omega_{1})\theta(k),
\end{equation}
where $\omega_{1}$ is a smoothing coefficient hyperparameter, and set as 0.95 in this work. $\theta'(k)$ and $\theta(k)$ denote the weights at the $k$-th iteration for the teacher and student models, respectively. 

The seen samples $\mathcal{D}_s$ are fed into the student network and classification network for visual embedding.
Therefore, the seen visual constraint $L_{svc}$ is defined as:
\begin{equation}
    \mathcal{L}_{svc} = - \displaystyle \mathop{\mathbb{E}}_{(x_s,y_s) \in D_s}[y_{s}^{\top} \ln C(S(x_{s}))].
\end{equation}
where $S(\cdot)$ and $C(\cdot)$ denotes the student network and classification network.

Since labels for unseen images $\mathcal{X}_{u}$ cannot access, how to apply unsupervised representation learning becomes the critical component for unbiased visual embedding.
Recently, data augmentation has been proved to be an effective way for unsupervised or semi-supervised representation learning.
Using the data augmentation can construct an unsupervised constraint between the unlabeled image and its augmentations, which can be used to infer the visual representation for unlabeled images.

Given the unseen images $\mathcal{X}_{u}$, we firstly generate several augmented samples for each unseen image.
The unseen images $x_{t}$ along with their augmented samples $\hat{x}_{t}$ are fed into the student network $S$ and the teacher network $T$, whose embedded features are denoted as $S(x_t)$ and $T(\hat{x}_t)$, respectively.
Since $S(x_t)$ and $T(\hat{x}_t)$ are both the description for the target image $x_t$, $S(x_t)$ and $T(\hat{x}_t)$ should have a small difference.
Therefore, the consistency loss $L_{uvc}$ between $S(x_t)$ and $T(\hat{x}_{t})$ is measured by the Euclidean distance:
\begin{equation}
    \mathcal{L}_{uvc}= \displaystyle \mathop{\mathbb{E}}_{x_{t} \in D_t}[||S(x_t) - T(\hat{x}_{t})||^2].
\end{equation}

The total loss $L_{v}$ of the unbiased visual embedding is the sum of the seen visual constraint and unseen visual constraint:
\begin{equation}
    L_{v} = L_{svc} + \omega_{2}L_{uvc},
\end{equation}
where $\omega_{2}$, which is computed based on the iteration epoch number, is an unsupervised loss weighting function introduced in ~\cite{laine2016temporal}. 

Once obtaining the unbiased visual embedding, the visual embedding $\mathbf{v}_{i}$ for image $x_{i}$ is obtained by fedding the images into student network:
\begin{equation}
\mathbf{v}_{i}=S(x_{i}),
\end{equation}
where $\mathbf{v}_{i}\in \mathbb{R}^{1\times 2048}$. 
The visual embedding for seen and unseen domains are denoted as $\mathbf{V}_{s}$ and $\mathbf{V}_{u}$, respctively.

\subsection{Attentional Graph Attribute Embedding}
\begin{figure}
	\begin{center}
		\includegraphics[width=0.8\linewidth]{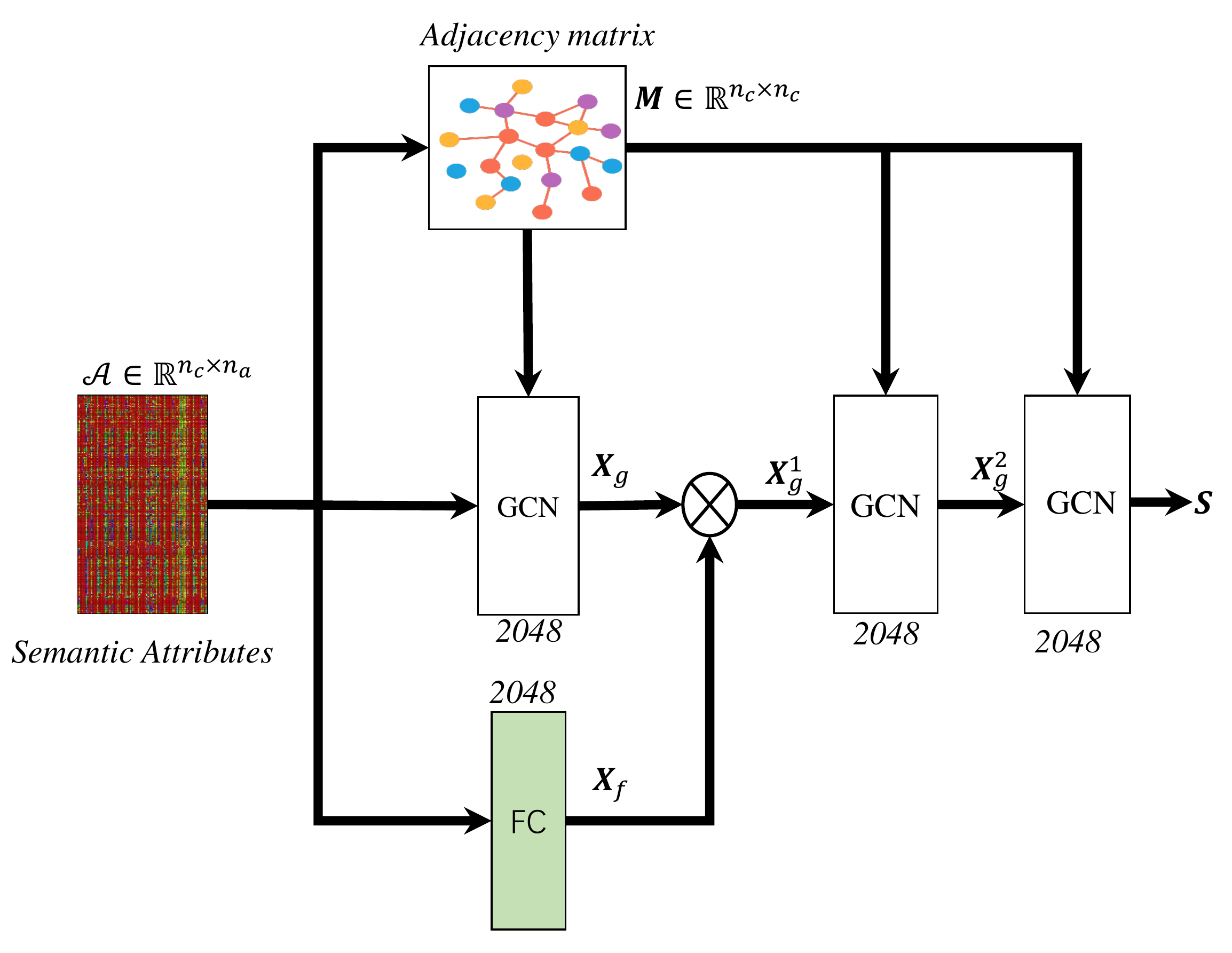}
	\end{center}
	\caption{The framework of Attentional Graph Attribute Embedding. The given semantic attributes $\mathcal{A}$ are used to generate the category adjacency matrix $\mathbf{M}$, the graph-based semantic embedding $\mathbf{X}_{g}$, and the category-independently semantic embedding $\mathbf{X}_{f}$. The $\mathbf{X}^{1}_{g}$ is obtainded by fusing the $\mathbf{X}_{g}$ and $\mathbf{X}_{f}$. The two following graph operations are applied to transfer the $\mathbf{X}^{1}_{g}$ to the semantic attribute embedding $\mathbf{S}$.}
	\label{fig3}
\end{figure}

Given the semantic attributes $\mathcal{A}=\{\mathcal{A}_{1},\mathcal{A}_{2},......,\mathcal{A}_{n_{c}}\}$ for all categories, where $\mathcal{A}_{i}\in R^{1\times n_{a}}$ and $n_{c}=n_{s}+n_{u}$ denotes the number of categories, the attribute embedding module $\mathcal{E}$ aims to map the attributes $\mathcal{A}$ into the semantic space $\mathbf{S}=\mathcal{E}(\mathcal{A})$.
Note that the $n_{a}, n_{c}, n_{s}$, and $n_{u}$ denotes the number of attributes, categories, seen categories, and unseen categories.
Once transferring the semantic attributes to the semantic space $\mathbf{S}$, the ZSL becomes the nearest neighbor (NN) search problem between the visual feature $\mathbf{V}$ and semantic space $\mathbf{S}$.
As introduced above, a good semantic embedding should make the seen and unseen semantic space have a minor bias. 
Therefore, we propose an Attentional Graph Attribute Embedding model by exploiting the semantic relationships between categories to reduce the semantic bias between seen and unseen domains, as shown in Figure~\ref{fig3}.

By taking the relationship between all categories as a structured graph, we apply the graph operation to embed the relationship between all categories.
A graph operation is implemented based on the normalized adjacency matrix $\mathbf{M}$ and the input feature map $\mathbf{X}$, where the normalized adjacency matrix $\mathbf{M}$ is computed with Eq.~\eqref{Eq:S}:
\begin{equation}
\mathbf{M}=\Lambda^{-\frac{1}{2}}(\mathbf{R}+\mathbf{I})\Lambda^{-\frac{1}{2}},
\label{Eq:S}
\end{equation}
where $\mathbf{I}$ is an identity matrix that represents self-connection, and $\mathbf{R}=am(A)$ is the adjacency matrix that denotes the similarity among all categories.
Once obtaining the normalized adjacency matrix $\mathbf{M}$, the graph operation is defined with the following formula:
\begin{equation}
\mathbf{X}_{out}=\mathbf{M}\mathbf{X}^{\top}\mathbf{W},
\end{equation}
where $\mathbf{W}$ denotes the trainable weight matrix.

With the graph operation, we propose an Attentional Graph Attribute Embedding model for semantic embedding.
Given the semantic attributes $\mathcal{A}$ and the normalized adjacency matrix $\mathbf{M}$, we firstly obtain the graph-based semantic embedding $\mathbf{X}_{g}$ by,
\begin{equation}
\mathbf{X}_{g}=\mathbf{M}\mathcal{A}\mathbf{W}_{g},
\end{equation}
where $\mathcal{A}$ is the input attribute, and $\mathbf{W}_{g}$ is the corresponding weight. 

Besides the graph-based semantic embedding, we also employ a category-independent semantic embedding, which is complementary to the graph-based semantic embedding.
The category-independent semantic embedding is obtained by applying a fully-connected network to transfer the source semantic attributes $\mathcal{A}$ into the corresponding semantic feature $\mathbf{X}_{f}$,
\begin{equation}
\mathbf{X}_{f}=\sigma(\mathcal{A}\mathbf{W}_{f}),
\end{equation}
where $\sigma(\cdot)$ denotes the non-linear operation, and $\mathbf{W}_{f}$ is the weight.

Since the semantic feature $\mathbf{W}_{f}$ have different characteristics from the graph semantic feature $\mathbf{W}_{g}$, fusing those two features can boost the semantic embedding,
\begin{equation}
\mathbf{X}^{1}_{g}=\sigma(\mathbf{X}_{g}\mathbf{X}_{f}).
\end{equation}

After obtaining the attentional-based semantic embedding $\mathbf{X}^{1}_{g}$, two graph operations are used to embed the semantic information further, and the final semantic embedding is denoted as $\mathbf{S}$,
\begin{equation}
\mathbf{X}^{2}_{g}=\sigma(\mathbf{M}\mathbf{X}^{1}_{g}\mathbf{W}^{2}_{g}),
\end{equation}
\begin{equation}
\mathbf{S}=\sigma(\mathbf{M}\mathbf{X}^{2}_{g}\mathbf{W}^{3}_{g}).
\end{equation}
where $\mathbf{W}^{2}_{g}$ and $\mathbf{W}^{3}_{g}$ are the corresponding weights.

\subsection{Semantic-Visual Alignment}
After obtaining the unbiased semantic embedding and visual embedding module, we then explore how to associate the visual and semantic embedding spaces in two domains.
There are two ways to associate the visual and semantic spaces: 1) fixing the visual space and optimizing the semantic embedding; and 2) fixing the semantic space and optimizing the visual embedding.
Compared with the semantic space, the visual space is more discriminative due to the supervised representation learning.
Therefore, we first fix the visual space learned by Unbiased Visual Embedding, and then optimize the Attentional Graph Attribute Embedding to be aligned with the fixed visual space via the proposed Semantic-Visual Alignment constraint.

Since the image-label relationship is given in the seen domain, the euclidean distance is used to minimize the distance between the visual image features and corresponding semantic category attributes.
Since images belonging to each category have large inter-class differences, using the image visual feature to align the semantic representation may cause a large bias between visual and semantic spaces. 
Therefore, the category representation, which is the mean of all visual features belonging to the same category, has a minor bias with the category semantic space.
As a consequence, the Seen Semantic Alignment constraint is the euclidean distance between the category visual representation and semantic embedding:
\begin{equation}
\mathcal{L}_{ssa}=\frac{1}{n_{s}}\sum_{i=1}^{n_{s}}||\frac{1}{n_{s}^{i}}\sum_{x\in \mathcal{D}_{s}^{i}}S(x)-\mathcal{E}(\mathcal{A}_{i})||^{2},
\end{equation}
where $n_{s}$ and $n_{s}^{i}$ are the number of the seen categories and number of images belong to the $i$-th seen category.
$\mathcal{D}_{s}^{i}$ denotes the images for $i$-th seen category.
$\mathcal{E}(\cdot)$ is the proposed Attentional Graph Attribute Embedding model and $\mathcal{A}_{i}$ is the attribute for $i$-th category.

For the unseen images, the Unseen Semantic Alignment (USA) constraint is used to associate their visual and semantic spaces.
Since there is no ground-truth label for unseen images, we minimize the euclidean distance between each unseen visual feature and its nearest neighbor among all unseen semantic space,
\begin{equation}
\mathcal{L}_{usa}=\frac{1}{N}\sum_{i=1}^{N}\min_{j\in[1,n_{u}]}||S(x_{u}^{i})-\mathcal{E}(\mathcal{A}_{j})||^{2},
\label{eq:usva}
\end{equation}
where $x_{u}^{i}$ is the $i$-th unseen images, and $S(\cdot)$ is the student network. $N$ and $n_{u}$ is the number of unseen images and unseen categories.

Since the visual feature $S(x_{u}^{i})\in \mathbb{R}^{1\times 2048}$ and semantic features $\mathcal{E}(\mathcal{A}_{u})\in \mathbb{R}^{n_u\times 2048}$ are both L2-normalized, the $S(x_{u}^{i})\mathcal{E}(\mathcal{A}_{u})^{\top}$ can represent its cosine similairty, \emph{e.g.,} the higher score, the higher similarity.
Finally, Eq.~\eqref{eq:usva} can be reformulated as:
\begin{equation}
\mathcal{L}_{usa}=\frac{1}{N}\sum_{i=1}^{N}\{1-\max[S(x_{u}^{i})\mathcal{E}(\mathcal{A}_{u})^\top]\},
\end{equation}
where $\mathcal{A}_{u}$ is the semantic attributes for all unseen categories.

The final loss for the semantic embedding is $\mathcal{L}_{agae}$:
\begin{equation}
\mathcal{L}_{agae}=\mathcal{L}_{ssa}+\omega_{3}\mathcal{L}_{usa},
\end{equation}
where $\omega_{3}$ is the weight to balance the effect of seen and unseen semantic alignment constraints.

\section{Experiments}
\subsection{Datasets and Evaluation Metrics}
\paragraph{Datasets} To evaluate the effectiveness of the  proposed Attentional Graph Attribute Embedding module, we perform several evaluations on three ZSL datasets, \emph{i.e.,} Caltech-USCD Birds-200-2011 (CUB) \cite{wah2011caltech}, SUN \cite{patterson2012sun}, and Animals with Attributes2 (AWA2) \cite{xian2018zero}.
Similar to the existing work~\cite{wan2019transductive}, the evaluations are performed under two different data split strategies: 1) Standard Splits (SS) that is firstly presented in ~\cite{lampert2009learning} and has been widely used for ZSL. 2) Proposed Splits (PS): since the unseen domain of Standard Splits (SS) often has some overlap category with ImageNet dataset, PS presents a newly and reasonably seen/unseen class split.
The details Standard Splits (SS)  and Proposed Splits (PS)~\cite{xian2018zero}  about each dataset are shown in Table~\ref{tab:data_ss} and Table~\ref{tab:data_ps}.
\begin{table}
\begin{center}
\caption{The details for each dataset under Standard Splits (SS). } \label{tab:data_ss}
\begin{tabular}{lccccc}
  \hline
  Datasets &Attributes&$|\mathcal{Y}_s|$&$|\mathcal{Y}_u|$&TrainVal&Test \\
  \hline
  \hline
  CUB~\cite{wah2011caltech}&312&150&50&8,855&2,933\\
  AWA2~\cite{xian2018zero}&85&40&10&30,337&6,985\\
  SUN~\cite{patterson2012sun}&102&645&72&13,339&1,492\\
  \hline
\end{tabular}
\end{center}
\end{table}

\begin{table}
\begin{center}
\caption{The details for each dataset under proposed splits (PS). } \label{tab:data_ps}
\begin{tabular}{|l|c|c|c|c|c|c|}
  \hline
  \multirow{2}*{Datasets}&\multirow{2}*{Attributes}&\multirow{2}*{$|\mathcal{Y}_s|$}&\multirow{2}*{$|\mathcal{Y}_u|$}&Train&\multicolumn{2}{|c|}{Test} \\
  \cline{5-7} 
                 &                  &                             &                                &Seen&Seen&Unseen \\
  \hline
  \hline
  CUB~\cite{wah2011caltech}&312&150&50&7,057&1,764&2,967\\
  AWA2~\cite{xian2018zero}&85&40&10&23,527&5,882&7,913\\
  SUN~\cite{patterson2012sun}&102&645&72&10,320&2,580&1,440\\
  \hline
\end{tabular}
\end{center}
\end{table}

\paragraph{Evaluation Metrics} 
Based on whether considering the images belonging to the seen category during the evaluation, there exist two different evaluation settings.
The first one is Conventional ZSL setting, which only recognizes the images belongs to the unseen category.
In order to better evaluate the generalization and robustness of the algorithm, the Generalized ZSL, where test images come from both seen and unseen categories, has been widely used recently.
Similar to previous methods, the Multi-way Classification Accuracy (MCA) is used in the Conventional ZSL setting.
For the Generalized ZSL, the harmonic mean, denoted by  $H = (2\times MCA_u\times MCA_s)/(MCA_u+MCA_s)$, is used to evaluate the comprehensive performance.  $MCA_s$ and $MCA_u$ are the Multi-way Classification Accuracy ($MCA$) for seen and unseen domains, respectively.

\subsection{Implementation Details}
The proposed method is implemented with the Pytorch framework based on the existing work~\cite{wan2019transductive}\footnote{https://github.com/raywzy/VSC}.
For the visual embedding, the student network in Mean-Teacher framework is adopted based on the pretrained ResNet-101~\cite{he2016deep}, and data augmentations, \emph{i.e.,} horizontally flip and randomly crop,  are used to increase data diversity for inferring the unseen visual representations.
The hidden unit number of layers in graph embedding networks are both 2048 besides the input layer, which depends on the number of semantic attributes.
When align and measure the visual and semantic spaces, the visual and semantic features are both L2-normalized.
For the visual embedding, we apply Adam optimizer with the learning rate of 0.001 to train the mean-teacher networks. 
Since the number of categories are different and each category contains a different number of images, we set the different training setting for each dataset.
During semantic-visual alignment, the hyperparameter $\omega_{3}$ is set as 0.001, 0.01, and 0.001 for CUB, AwA2, and SUN datasets, respectively.

\subsection{Ablation Studies}
The contributions of the proposed Attribute-Induced Bias Eliminating (AIBE) are to introduces three constraints to reduce three bias of Transductive ZSL: 1) using unseen visual constraint to reduce the \emph{visual bias} between seen and unseen domains; 2) using graph attribute embedding to reduce the \emph{semantic bias} between seen and unseen domains; 3) using unseen semantic-visual alignment to reduce the \emph{unseen bias} between visual and semantic spaces.
We thus analyze the effectiveness of each constraint.

\begin{figure}
	\begin{center}
		\includegraphics[width=0.8\linewidth]{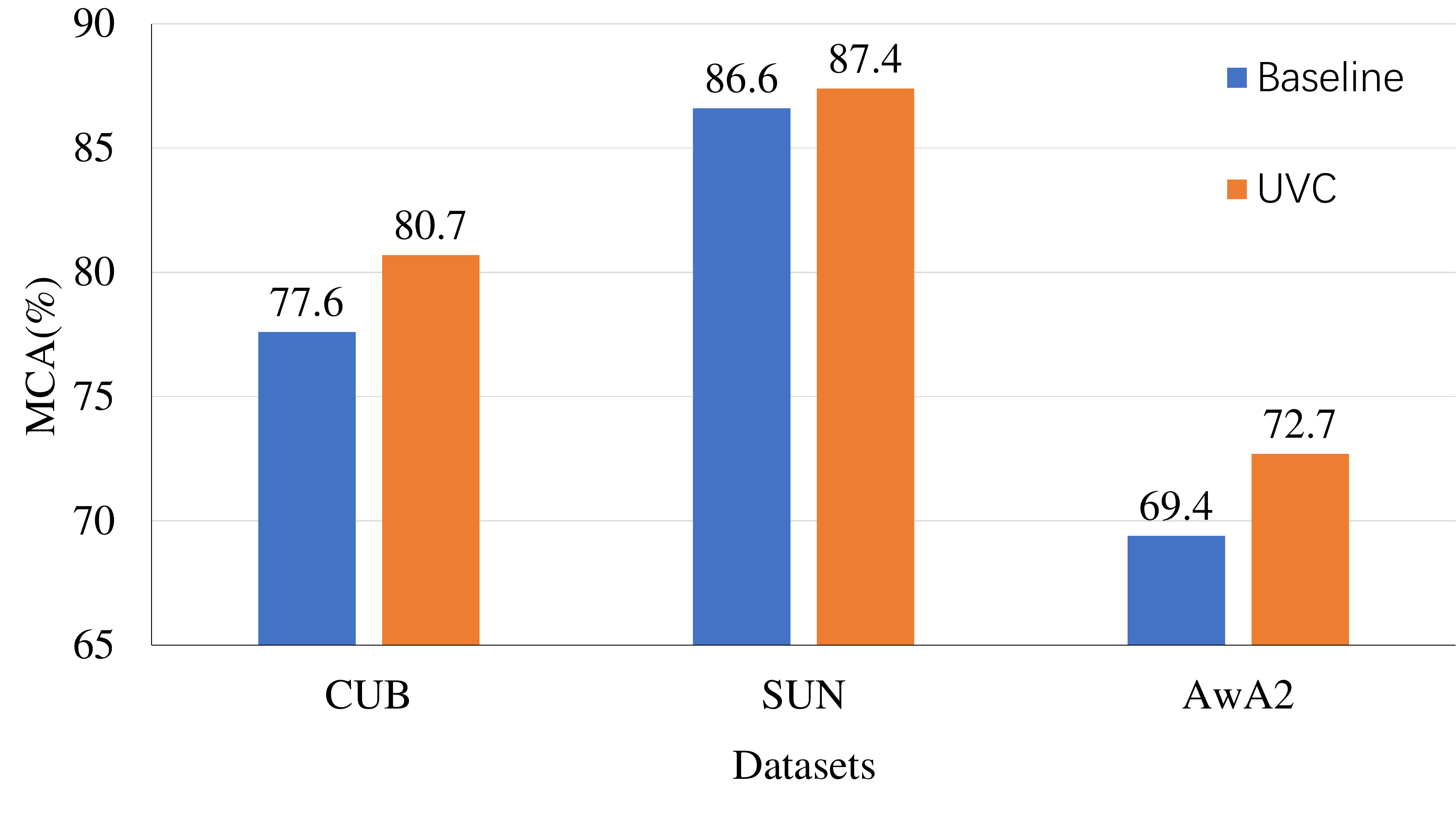}
	\end{center}
	\caption{Effect of Unseen Visual Constraint (UVC) for Conventional ZSL under Proposed Split (PS) setting. The Baseline model is inferred only based on Seen Visual Constraint, and `UVC' denotes the model by combining the Seen Visual Constraint and Unseen Visual Constraint.}
	\label{fig4}
\end{figure}

\noindent\textbf{Effect of unseen visual constraint}
To reduce the visual bias between seen and unseen domains, we treat the unsupervised Mean-Teacher model as a visual embedding model by considering the unseen visual constraint $L_{uvc}$.
Therefore, we firstly analyze the effect of unseen visual constraint and summarize the related results  in Figure~\ref{fig4}.
In Figure~\ref{fig4}, the `Baseline' denotes that only the seen visual constraint $L_{svc}$ is used to infer the visual embedding module, and `UVC' denotes combining the seen visual constraint $L_{svc}$ and unseen visual constraint $L_{uvc}$.
As shown in Figure~\ref{fig4}, using the unseen visual constraint obtains the consistent improvement upon Conventional ZSL under Proposed Split (PS) setting for all three datasets, \emph{e.g.}, improving the MCA from 77.6\%, 86.6\%, and 69.4\% to 80.7\%, 87.4\% and 72.7\% for CUB, AwA2, and SUN, respectively.
The improvement demonstrates that using the unseen visual constraint can increase the distinguishability of visual embeddings, and reduce the visual bias between seen and unseen domains in Transductive ZSL.

\noindent\textbf{Effect of Attentional Graph Attribute Embedding}
We also evaluate the effect of Attentional Graph Attribute Embedding (AGAE) for semantic space learning, and summarize the detailed results in Table~\ref{tab:effect_agae}.
By discarding the relationship among all categories, we provide a three fully-connected embedding (FCE) models.
Besides the FCE model, we also define a Graph Attribute Embedding (GAE) model, which is implemented with three graph layers and discards category-independently semantic representation.
To make a fair comparison with the AGAE, each layer of FCE and GAE consists of 2,048 neurons beside the input layer.
As shown in Table~\ref{tab:effect_agae}, the Graph Attribute Embedding (GAE) is superior to the fully-connected embedding (FCE), \emph{e.g.,} improving the MCA from 74.1\% and 74.4\% to 79.1\% and 76.4\% for SS and PS settings on CUB dataset, respectively.
The larger improvement proves the advantage of considering the relationship between categories during semantic embedding.
We further compare the Graph Attribute Embedding (GAE) and Attentional Graph Attribute Embedding (AGAE).
Since the GAE has been achieved better performance for AwA2 under the SS setting, the AGAE achieves the same performance as GAE.
In addition to the AwA2 dataset, the proposed AGAE obtains a higher performance than GAE on the rest settings.
The higher performance proves that treating the category-independently semantic representation as an attentional semantic space can boost the semantic space generated by GAE.
Therefore, we can conclude that the proposed Attentional Graph Attribute Embedding is an effective way for semantic attribute embedding in Zero-shot learning.

\begin{table}
\begin{center}
\caption{Effect of Attentional Graph Attribute Embedding for Conventional ZSL.}\label{tab:effect_agae}
\begin{tabular}{c|ccc|ccc}
  \hline
  Settings & \multicolumn{3}{|c|}{SS}  &\multicolumn{3}{|c}{PS} \\
\hline
  Methods &  CUB & AsA2 & SUN & CUB & AwA2 & SUN \\
\hline
  \hline
FCE &74.1&96.0&71.7&74.4&64.9&72.0\\
GAE & 79.1&97.1&72.4&76.4&77.6&72.2\\
AGAE & 82.8&97.1&73.2&80.7&87.4&72.7\\
\hline
\end{tabular}
\end{center}
\end{table}

\begin{figure*}
\begin{center}
\subfigure[CUB: Without USA]{
\includegraphics[width=0.28\linewidth]{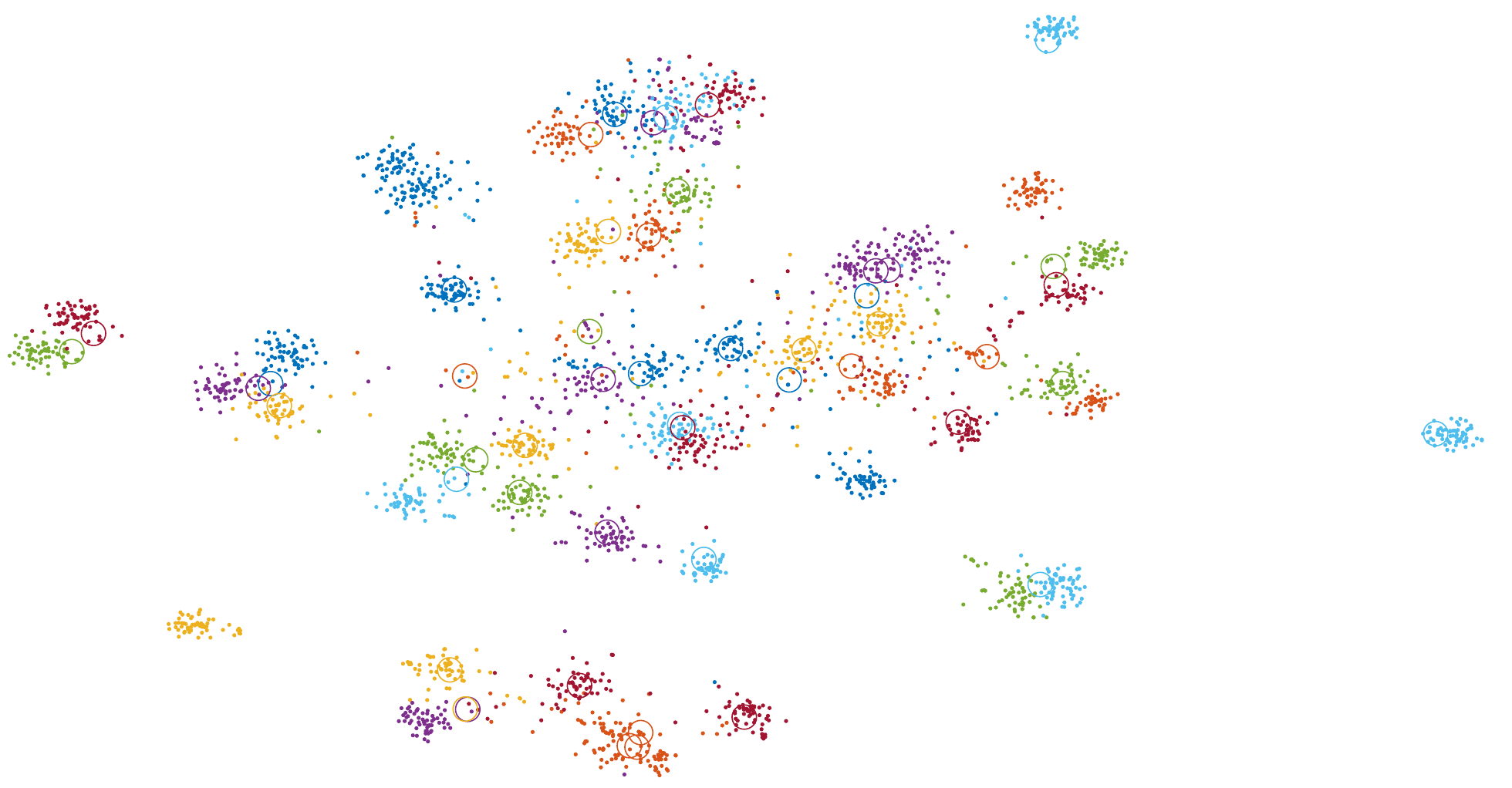}}
\subfigure[AwA2: Without USA]{
\includegraphics[width=0.28\linewidth]{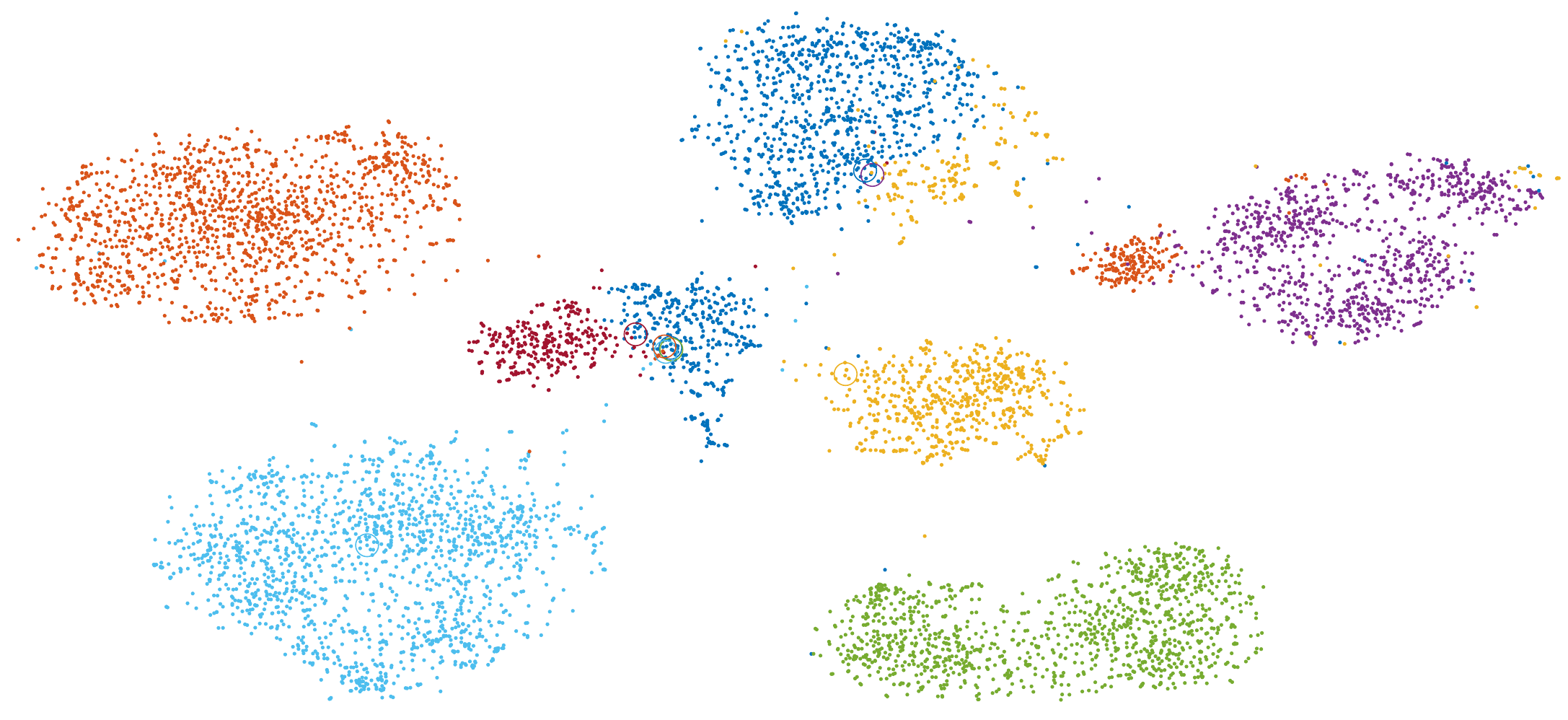}}
\subfigure[SUN: Without USA]{
\includegraphics[width=0.28\linewidth]{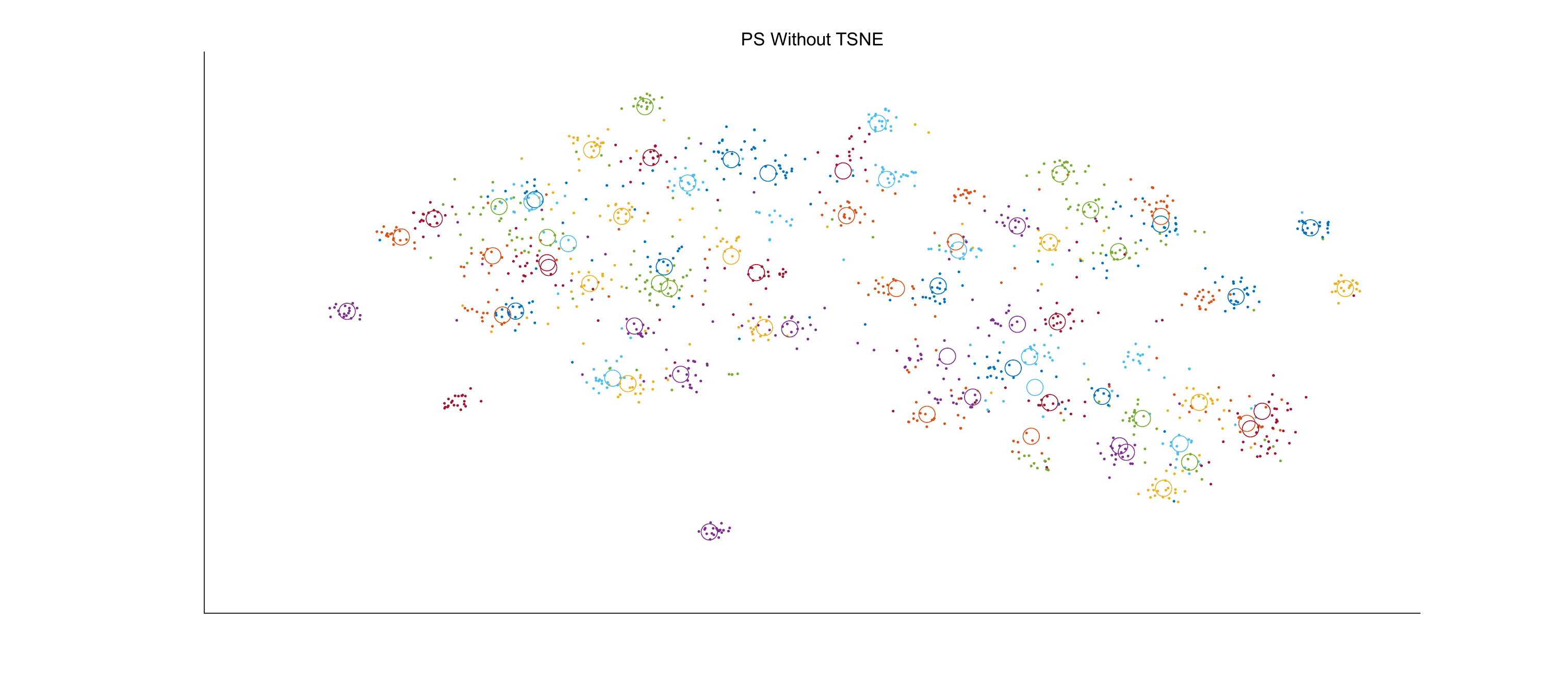}}\\
\subfigure[CUB: With USA]{
\includegraphics[width=0.28\linewidth]{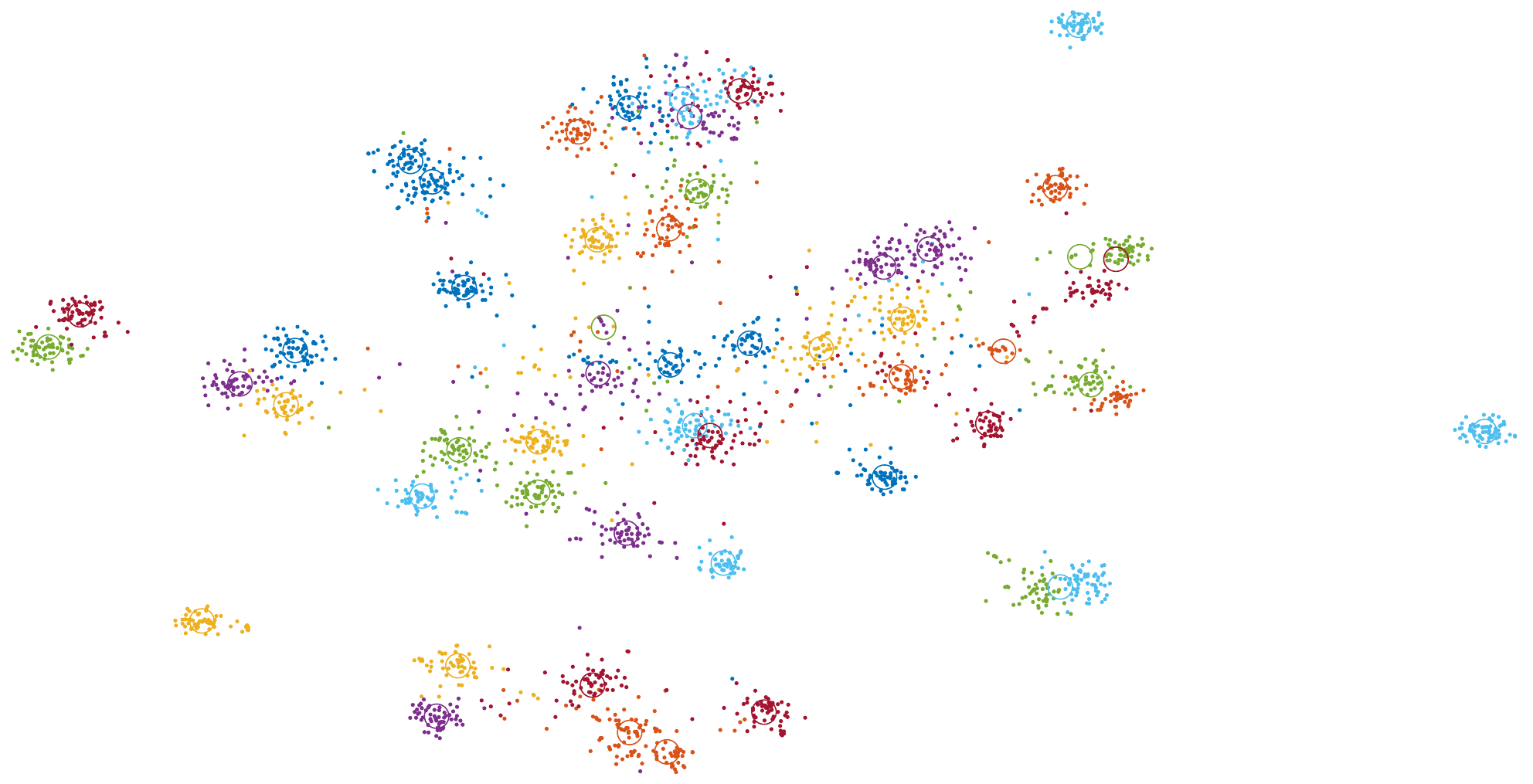}}
\subfigure[AwA2: With USA]{
\includegraphics[width=0.28\linewidth]{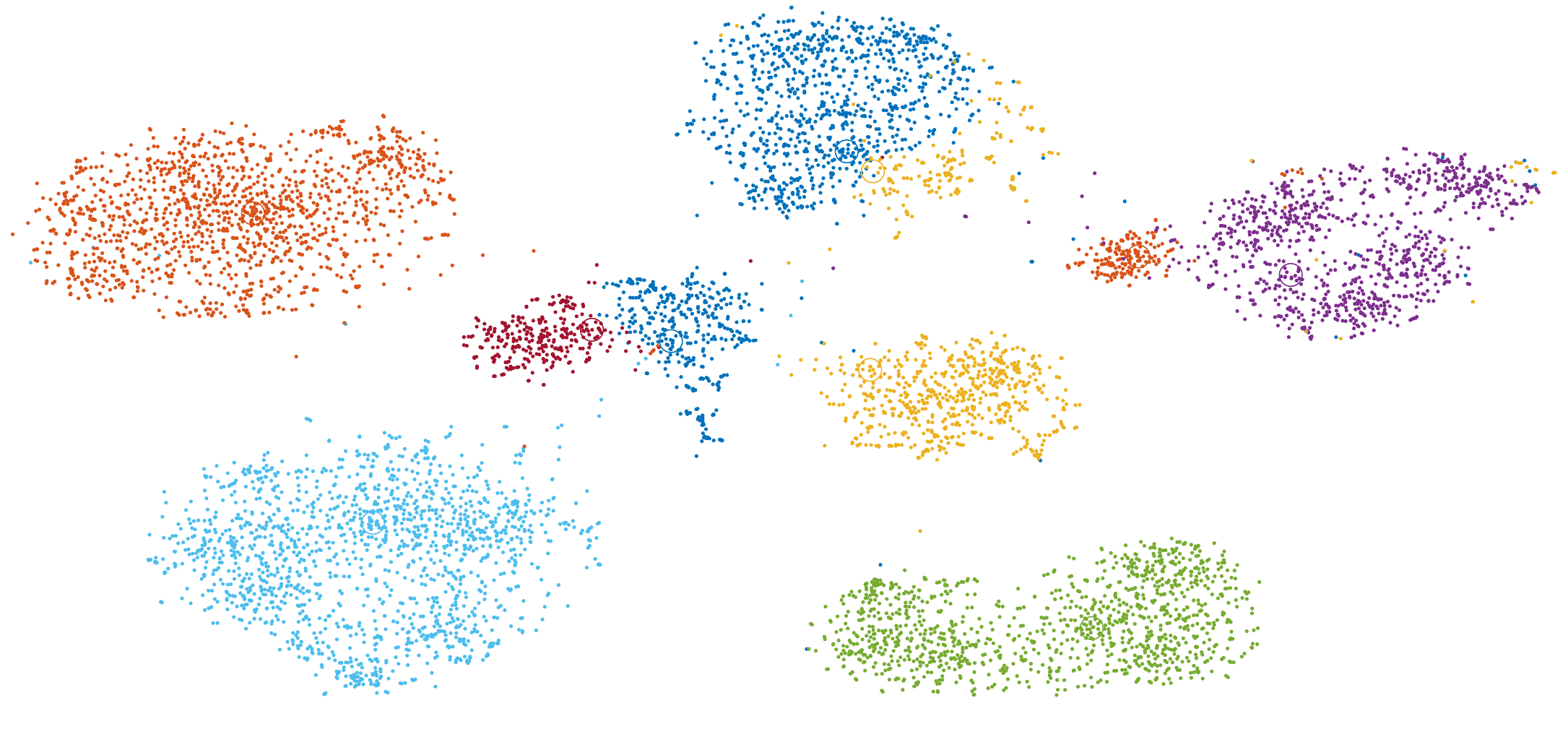}}
\subfigure[SUN: With USA]{
\includegraphics[width=0.28\linewidth]{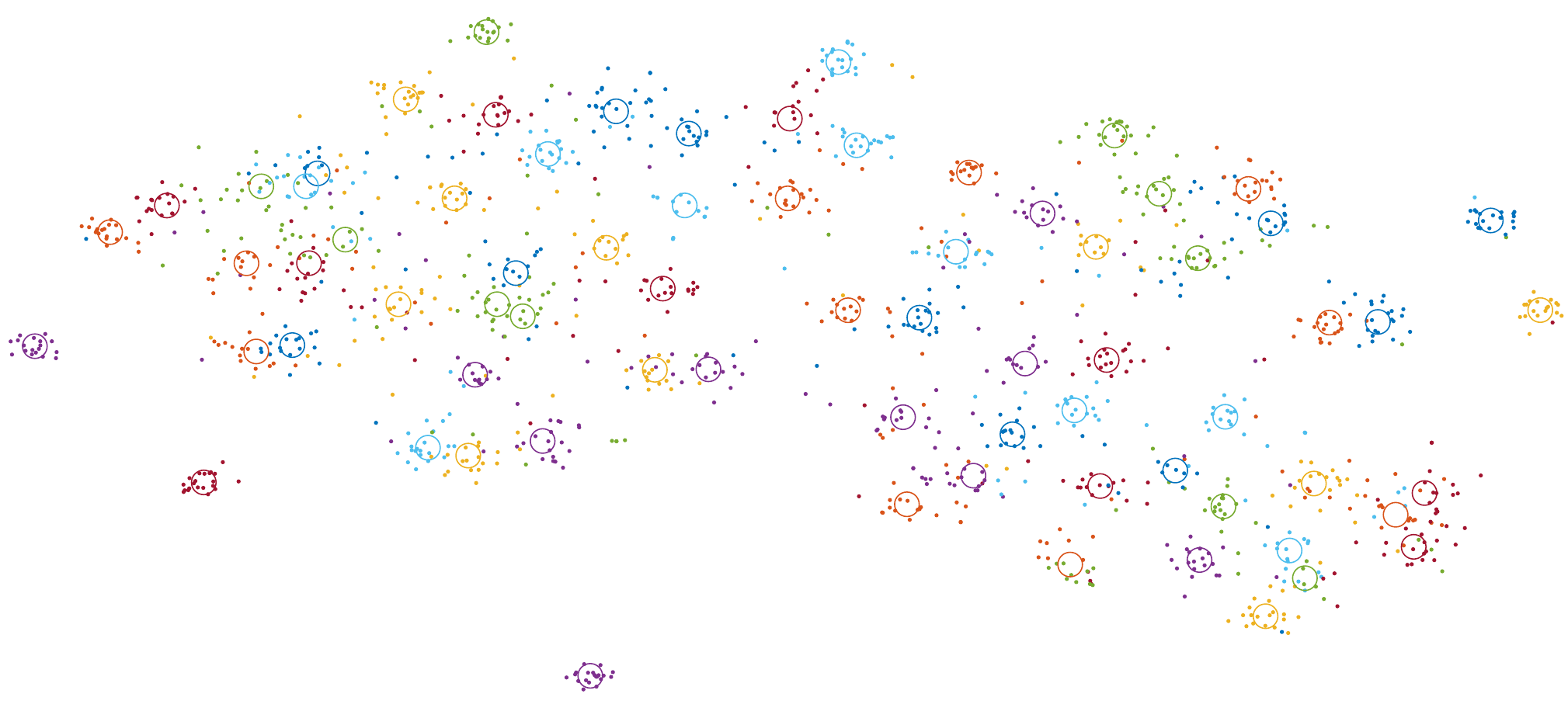}}
\end{center}
\caption{Visualization of visual and semantic embedding distribution of unseen classes using t-SNE~\cite{maaten2008visualizing}. The point ``$\bullet$" denotes the generated visual description for each unseen images, and the cycle ``$\bigcirc$" represents the generated semantic embedding for each category. Without and with USA denote whether using the Unseen Semantic Alignment (USA) constraint during semantic embedding. It can be seen that the semantic centers generated by using USA constraint are distinguishable}
\label{Fig:tsne}
\end{figure*}

\noindent\textbf{Effect of Unseen Semantic Alignment}
To reduce the bias between visual and semantic spaces for the unseen domain, we propose an Unseen Semantic Alignment (USA) constraint to associate the unseen semantic and visual spaces.
We thus evaluate the effect of USA constraint, and summarize the results in Table~\ref{tab:effect_usa}.
Based on the visual and semantic features generated with the Unbiased Visual Embedding and the Attentional Graph Attribute Embedding, the SSC and AGAE model are implemented without and with using the USA constraint.
\begin{table}
\centering
\caption{Effect of Unseen Semantic Alignment for Conventional ZSL.}\label{tab:effect_usa}
\begin{tabular}{l|ccc|ccc}
  \hline
  Settings & \multicolumn{3}{|c|}{Conventional ZSL(SS)}  &\multicolumn{3}{|c}{Conventional ZSL(PS)} \\
 \hline
  Methods &  CUB & AwA2 & SUN & CUB & AwA2 & SUN \\
  \hline
 \hline
Baseline &70.8&86.5&67.1&70&63.0&66.59\\
UVSC &82.8&97.1&73.2&80.7&87.4&72.7\\
\hline
Improvement &8.0&10.6&6.1&10.7&24.4&6.1\\
\hline
\end{tabular}
\end{table}
It can observe that the AGAE model with USA constraint obtains an obvious improvement upon the SSC model, \emph{e.g.,} 12.8\% and 10.7\%, 10.6\% and 14.4\%, 6.1\% and 6.1\%  improvement on SS and PS settings for CUB, AwA2, and SUN datasets, respectively.
We further give some qualitative results in Figure~\ref{Fig:tsne}.
The first and second rows illustrate the results for SSC and AGAE models, respectively.
From the first row in Figure~\ref{Fig:tsne}, we can observe that the semantic centers generated by the SSC model have relatively small distances among different categories.
However, there is even much overlap between some semantic centers, which means that the semantic centers generated by the SSC model are easily to be confused.
Compared with the SSC model, we can see that the model AGAE with Unseen Semantic Alignment can increase the distance between different semantic centers, as shown in the second row in Figure~\ref{Fig:tsne}.
Therefore, the semantic centers generated by AGAE are more distinguishable.
The above analysis demonstrates that the proposed Unseen Semantic Alignment constraint is a powerful constraint to align the visual and semantic spaces for the unseen domain.
Furthermore, the Unseen Semantic Alignment constraint is independent of visual embedding and semantic embedding modules, thus it can be applied to any ZSL algorithm to boost the performance.
\begin{table}
\begin{center}
\caption{Comparision under Conventional ZSL setting.}\label{tab:czsl}
\begin{tabular}{c|cc|cc|cc}
  \hline
  Datasets & \multicolumn{2}{|c|}{CUB}  &\multicolumn{2}{|c}{AwA2} &\multicolumn{2}{|c}{SUN}\\
 \hline
  Methods &  SS & PS & SS & PS & SS & PS \\
  \hline
  SE-ZSL~\cite{kumar2018generalized} &  60.3 & 59.6 & 80.8 & 69.2 & 64.5 & 63.4 \\
  QFSL~\cite{song2018transductive} &  69.7 & 72.1& 84.8 & 79.7 & 61.7 & 58.3 \\
  WDVSc~\cite{wan2019transductive} &  74.2 & 73.4 & 96.7 & 87.3 & 67.8 & 63.4 \\
  $LFGAA^{R}$~\cite{liu2019attribute} &  79.7 & 78.9 & 94.4& 84.8& 64.0 & 66.2 \\
  GXE~\cite{li2019rethinking}&  - & 61.3& - & 83.2& - & 63.5 \\
  \hline
  \hline
  AIBE&  \textbf{82.8} & \textbf{80.7} & \textbf{97.1} & \textbf{87.4} & \textbf{73.2}& \textbf{72.7} \\
   \hline
\end{tabular}
\end{center}
\end{table}

\begin{table*}
\begin{center}
\caption{Comparision under Generalized ZSL setting. } \label{tab:gzsl}
\begin{tabular}{c|ccc|ccc|ccc}
  \hline
    & \multicolumn{3}{|c|}{CUB}  &\multicolumn{3}{|c}{AwA2} &\multicolumn{3}{|c}{SUN}\\
 \hline
  Methods &  $MCA_u$ & $MCA_{s}$ & H & $MCA_{u}$ & $MCA_{s}$ & H & $MCA_{u}$ & $MCA_{s}$ & H\\
  \hline
  \hline
  WDVSc~\cite{wan2019transductive} & 43.4 & \textbf{85.4} & 57.5 & 76.4 & 88.1 & 81.8 & & & \\
  GXE~\cite{li2019rethinking} & 57.0 & 68.7 & 62.3 & \textbf{80.2} & 90.0 & \textbf{84.8} & 45.4&\textbf{58.1}&51.0\\
  QFSL~\cite{song2018transductive} & 71.5 & 74.9 & 73.2 & 66.2 & 93.1 & 77.4 & 31.2&51.31&38.8\\
  DSRL~\cite{norouzi2013zero} & 17.3 & 39.0 & 24.0 & 20.8 & 74.7 & 32.6 & 17.7&25.0&20.7\\
   \hline
    \hline
   AIBE &\textbf{72.4} & 78.7& \textbf{75.5} & 75.4 & \textbf{90.7} & 82.5 & \textbf{56.1} & 47.5 & \textbf{52.1} \\
     \hline
\end{tabular}
\end{center}
\end{table*}

\subsection{Comparison with state-of-the-art methods}
In this section, we make the comparision with several state-of-the-art transductive ZSL methods under both Conventional Zero-shot learning and Generalized Zero-shot learning, and summarize the results in Table~\ref{tab:czsl} and Table~\ref{tab:gzsl}, respectively.
\emph{Note that we only make comparisons with the \textbf{Transductive ZSL} methods in this work, and not consider the Inductive ZSL methods}.
\paragraph{\textbf{Conventional Zero-shot learning}} The conventional Zero-shot learning, which only tests the images belonging to the unseen categories, is a widely used evaluation setting for ZSL.
We thus show the comparison with existing methods under Conventional ZSL in Table~\ref{tab:czsl} for both standard splits (SS) and proposed splits (PS) settings.
As shown in Table~\ref{tab:czsl}, by reducing the mentioned three bias,  the proposed method obtains better performance than existing methods on all datasets for SS and PS settings.
From Table~\ref{tab:czsl}, we also observe that all the methods except one on SUN dataset on the standard splits (SS) setting obtain a higher performance than the proposed splits (PS) setting due to that the unseen categories of standard splits (SS) setting have some overlap with the Imagenet.

Among all three datasets, the SUN dataset is the most challenging dataset because it contains a large number of categories, and each category contains few images.
Therefore, existing methods all obtain a worse performance on the SUN dataset, \emph{e.g.,} the existing state-of-the-art performance only ahcieves 67.8\%~\cite{wan2019transductive} and 66.2\%~\cite{liu2019attribute} for SS and PS settings.
Compared with the existing methods, the proposed AIBE obtains the MCA of 73.2\% and 72.7\% for SS and PS settings, respectively, which achieves the  5.4\% and 6.5\% improvements upon existing best performance.
The substantial improvement can demonstrate the effectiveness of the proposed  Attribute-Induced Bias Eliminating for zero-shot learning.

\paragraph{\textbf{Generalized Zero-shot learning}}
Besides the Conventional zero-shot learning, we also compare the proposed AIBE with existing methods under Generalized Zero-shot learning (GZSL) setting, and summarize the results in Table~\ref{tab:gzsl}.
Note that the Generalized Zero-shot learning assumes that the testing images come from both seen and unseen categories.
It can be seen that the proposed module still outperforms the existing methods in most cases. 
The only exception is that the GXE~\cite{li2019rethinking} is slightly higher than our method on the AwA2 dataset as the GXE pays much attention to domain recognization between seen and unseen domains.
Different from~\cite{li2019rethinking}, the AIBE focuses on reducing the bias for unseen domain and ignores domain recognization, which plays a critical role in Generalized ZSL.
By discarding the domain recognization, the comparisons on SS and PS settings for the only unseen domain have been demonstrated that the proposed AIBE is superior to GXE.
From Table~\ref{tab:gzsl}, we also observe that the proposed AIBE obtains an obvious improvement upon the existing methods for the $MCA_u$ metric, which is consistent with our research motivation.

\section{Conclusion}
Transductive Zero-shot learning aims to associate the visual and semantic spaces used for recognizing the unseen objects.
Existing work focuses on reducing the \emph{seen bias} for seen domain and ignores the other critical biases that are all related to unseen domain.
Therefore, we propose an Attribute-Induced Bias Eliminating for semantic attribute embedding.
To obtain an unbiased visual embedding, we apply the Mean-Teacher (MT) for representation learning by considering the labeled seen images and unlabeld unseen images simultaneously.
For the semantic attribute embedding, the Attentional Graph Attribute Embedding(AGAE), which can reduce the semantic bias between seen and unseen categories, is proposed by leveraging the graph operation to embed the attribute relationship between all categories.
To associate the visual and semantic spaces for unseen domain, an Unseen Visual-Semantic Alignment (UVSA) constraint is further proposed to reduce the bias between visual and semantic spaces.
The evaluation of several benchmarks demonstrates that the proposed module can effectively reduce the domain bias for ZSL.

As discussed in the above comparison, the proposed method performs very well in Conventional Zero-shot learning settings.
However, it might not be suitable for the Generalized Zero-shot learning due to not considering the domain recognition problem.
In future work, we will focus on improving the domain division during  semantic embedding for Generalized Zero-shot learning.

\newpage
\bibliographystyle{abbrv}
\bibliography{main}

\begin{thebibliography}{10}

\bibitem{Annadani2018}
Y.~Annadani and S.~Biswas.
\newblock Preserving semantic relations for zero-shot learning.
\newblock In {\em Proceedings of the IEEE Conference on Computer Vision and
  Pattern Recognition}, pages 7603--7612, 2018.

\bibitem{changpinyo2020classifier}
S.~Changpinyo, W.-L. Chao, B.~Gong, and F.~Sha.
\newblock Classifier and exemplar synthesis for zero-shot learning.
\newblock {\em International Journal of Computer Vision}, 128(1):166--201,
  2020.

\bibitem{french2017self}
G.~French, M.~Mackiewicz, and M.~Fisher.
\newblock Self-ensembling for visual domain adaptation.
\newblock {\em arXiv preprint arXiv:1706.05208}, 2017.

\bibitem{fu2015transductive}
Y.~Fu, T.~M. Hospedales, T.~Xiang, and S.~Gong.
\newblock Transductive multi-view zero-shot learning.
\newblock {\em IEEE transactions on pattern analysis and machine intelligence},
  37(11):2332--2345, 2015.

\bibitem{guo2017zero}
Y.~Guo, G.~Ding, J.~Han, and Y.~Gao.
\newblock Zero-shot recognition via direct classifier learning with transferred
  samples and pseudo labels.
\newblock In {\em Thirty-First AAAI Conference on Artificial Intelligence},
  2017.

\bibitem{he2016deep}
K.~He, X.~Zhang, S.~Ren, and J.~Sun.
\newblock Deep residual learning for image recognition.
\newblock In {\em Proceedings of the IEEE conference on computer vision and
  pattern recognition}, pages 770--778, 2016.

\bibitem{hubert2017learning}
Y.-H. Hubert~Tsai, L.-K. Huang, and R.~Salakhutdinov.
\newblock Learning robust visual-semantic embeddings.
\newblock In {\em Proceedings of the IEEE International Conference on Computer
  Vision}, pages 3571--3580, 2017.

\bibitem{jia2019deep}
Z.~Jia, Z.~Zhang, L.~Wang, C.~Shan, and T.~Tan.
\newblock Deep unbiased embedding transfer for zero-shot learning.
\newblock {\em IEEE Transactions on Image Processing}, 29:1958--1971, 2019.

\bibitem{kampffmeyer2019rethinking}
M.~Kampffmeyer, Y.~Chen, X.~Liang, H.~Wang, Y.~Zhang, and E.~P. Xing.
\newblock Rethinking knowledge graph propagation for zero-shot learning.
\newblock In {\em Proceedings of the IEEE Conference on Computer Vision and
  Pattern Recognition}, pages 11487--11496, 2019.

\bibitem{kipf2016semi}
T.~N. Kipf and M.~Welling.
\newblock Semi-supervised classification with graph convolutional networks.
\newblock {\em arXiv preprint arXiv:1609.02907}, 2016.

\bibitem{kodirov2015unsupervised}
E.~Kodirov, T.~Xiang, Z.~Fu, and S.~Gong.
\newblock Unsupervised domain adaptation for zero-shot learning.
\newblock In {\em Proceedings of the IEEE international conference on computer
  vision}, pages 2452--2460, 2015.

\bibitem{kumar2018generalized}
V.~Kumar~Verma, G.~Arora, A.~Mishra, and P.~Rai.
\newblock Generalized zero-shot learning via synthesized examples.
\newblock In {\em Proceedings of the IEEE conference on computer vision and
  pattern recognition}, pages 4281--4289, 2018.

\bibitem{laine2016temporal}
S.~Laine and T.~Aila.
\newblock Temporal ensembling for semi-supervised learning.
\newblock {\em arXiv preprint arXiv:1610.02242}, 2016.

\bibitem{lampert2009learning}
C.~H. Lampert, H.~Nickisch, and S.~Harmeling.
\newblock Learning to detect unseen object classes by between-class attribute
  transfer.
\newblock In {\em 2009 IEEE Conference on Computer Vision and Pattern
  Recognition}, pages 951--958. IEEE, 2009.

\bibitem{li2019rethinking}
K.~Li, M.~R. Min, and Y.~Fu.
\newblock Rethinking zero-shot learning: A conditional visual classification
  perspective.
\newblock In {\em Proceedings of the IEEE International Conference on Computer
  Vision}, pages 3583--3592, 2019.

\bibitem{li2018discriminative}
Y.~Li, J.~Zhang, J.~Zhang, and K.~Huang.
\newblock Discriminative learning of latent features for zero-shot recognition.
\newblock In {\em Proceedings of the IEEE Conference on Computer Vision and
  Pattern Recognition}, pages 7463--7471, 2018.

\bibitem{liu2019attribute}
Y.~Liu, J.~Guo, D.~Cai, and X.~He.
\newblock Attribute attention for semantic disambiguation in zero-shot
  learning.
\newblock In {\em Proceedings of the IEEE International Conference on Computer
  Vision}, pages 6698--6707, 2019.

\bibitem{maaten2008visualizing}
L.~v.~d. Maaten and G.~Hinton.
\newblock Visualizing data using t-sne.
\newblock {\em Journal of machine learning research}, 9(Nov):2579--2605, 2008.

\bibitem{norouzi2013zero}
M.~Norouzi, T.~Mikolov, S.~Bengio, Y.~Singer, J.~Shlens, A.~Frome, G.~S.
  Corrado, and J.~Dean.
\newblock Zero-shot learning by convex combination of semantic embeddings.
\newblock {\em arXiv preprint arXiv:1312.5650}, 2013.

\bibitem{patterson2012sun}
G.~Patterson and J.~Hays.
\newblock Sun attribute database: Discovering, annotating, and recognizing
  scene attributes.
\newblock In {\em 2012 IEEE Conference on Computer Vision and Pattern
  Recognition}, pages 2751--2758. IEEE, 2012.

\bibitem{paul2019semantically}
A.~Paul, N.~C. Krishnan, and P.~Munjal.
\newblock Semantically aligned bias reducing zero shot learning.
\newblock In {\em Proceedings of the IEEE Conference on Computer Vision and
  Pattern Recognition}, pages 7056--7065, 2019.

\bibitem{Radovanovic2010}
M.~Radovanovi{\'c}, A.~Nanopoulos, and M.~Ivanovi{\'c}.
\newblock Hubs in space: Popular nearest neighbors in high-dimensional data.
\newblock {\em Journal of Machine Learning Research}, 11(Sep):2487--2531, 2010.

\bibitem{rohrbach2013transfer}
M.~Rohrbach, S.~Ebert, and B.~Schiele.
\newblock Transfer learning in a transductive setting.
\newblock In {\em Advances in neural information processing systems}, pages
  46--54, 2013.

\bibitem{romera2015embarrassingly}
B.~Romera-Paredes and P.~Torr.
\newblock An embarrassingly simple approach to zero-shot learning.
\newblock In {\em International Conference on Machine Learning}, pages
  2152--2161, 2015.

\bibitem{sariyildiz2019gradient}
M.~B. Sariyildiz and R.~G. Cinbis.
\newblock Gradient matching generative networks for zero-shot learning.
\newblock In {\em Proceedings of the IEEE Conference on Computer Vision and
  Pattern Recognition}, pages 2168--2178, 2019.

\bibitem{snell2017prototypical}
J.~Snell, K.~Swersky, and R.~Zemel.
\newblock Prototypical networks for few-shot learning.
\newblock In {\em Advances in neural information processing systems}, pages
  4077--4087, 2017.

\bibitem{song2018selective}
J.~Song, C.~Shen, J.~Lei, A.-X. Zeng, K.~Ou, D.~Tao, and M.~Song.
\newblock Selective zero-shot classification with augmented attributes.
\newblock In {\em Proceedings of the European Conference on Computer Vision
  (ECCV)}, pages 468--483, 2018.

\bibitem{song2018transductive}
J.~Song, C.~Shen, Y.~Yang, Y.~Liu, and M.~Song.
\newblock Transductive unbiased embedding for zero-shot learning.
\newblock In {\em Proceedings of the IEEE Conference on Computer Vision and
  Pattern Recognition}, pages 1024--1033, 2018.

\bibitem{tarvainen2017mean}
A.~Tarvainen and H.~Valpola.
\newblock Mean teachers are better role models: Weight-averaged consistency
  targets improve semi-supervised deep learning results.
\newblock In {\em Advances in neural information processing systems}, pages
  1195--1204, 2017.

\bibitem{Tomasev2014}
N.~Tomasev, M.~Radovanovic, D.~Mladenic, and M.~Ivanovic.
\newblock The role of hubness in clustering high-dimensional data.
\newblock {\em IEEE transactions on knowledge and data engineering},
  26(3):739--751, 2014.

\bibitem{verma2017simple}
V.~K. Verma and P.~Rai.
\newblock A simple exponential family framework for zero-shot learning.
\newblock In {\em Joint European Conference on Machine Learning and Knowledge
  Discovery in Databases}, pages 792--808. Springer, 2017.

\bibitem{wah2011caltech}
C.~Wah, S.~Branson, P.~Welinder, P.~Perona, and S.~Belongie.
\newblock The caltech-ucsd birds-200-2011 dataset.
\newblock 2011.

\bibitem{wan2019transductive}
Z.~Wan, D.~Chen, Y.~Li, X.~Yan, J.~Zhang, Y.~Yu, and J.~Liao.
\newblock Transductive zero-shot learning with visual structure constraint.
\newblock In {\em Advances in Neural Information Processing Systems}, pages
  9972--9982, 2019.

\bibitem{wang2019survey}
W.~Wang, V.~W. Zheng, H.~Yu, and C.~Miao.
\newblock A survey of zero-shot learning: Settings, methods, and applications.
\newblock {\em ACM Transactions on Intelligent Systems and Technology (TIST)},
  10(2):1--37, 2019.

\bibitem{xian2018zero}
Y.~Xian, C.~H. Lampert, B.~Schiele, and Z.~Akata.
\newblock Zero-shot learning---a comprehensive evaluation of the good, the bad
  and the ugly.
\newblock {\em IEEE transactions on pattern analysis and machine intelligence},
  41(9):2251--2265, 2018.

\bibitem{ye2019progressive}
M.~Ye and Y.~Guo.
\newblock Progressive ensemble networks for zero-shot recognition.
\newblock In {\em Proceedings of the IEEE Conference on Computer Vision and
  Pattern Recognition}, pages 11728--11736, 2019.

\bibitem{zhang2020towards}
L.~Zhang, P.~Wang, L.~Liu, C.~Shen, W.~Wei, Y.~Zhang, and A.~Van Den~Hengel.
\newblock Towards effective deep embedding for zero-shot learning.
\newblock {\em IEEE Transactions on Circuits and Systems for Video Technology},
  2020.

\bibitem{zhang2017learning}
L.~Zhang, T.~Xiang, and S.~Gong.
\newblock Learning a deep embedding model for zero-shot learning.
\newblock In {\em Proceedings of the IEEE Conference on Computer Vision and
  Pattern Recognition}, pages 2021--2030, 2017.

\bibitem{zhu2019generalized}
P.~Zhu, H.~Wang, and V.~Saligrama.
\newblock Generalized zero-shot recognition based on visually semantic
  embedding.
\newblock In {\em Proceedings of the IEEE Conference on Computer Vision and
  Pattern Recognition}, pages 2995--3003, 2019.

\end{thebibliography}

\end{document}